\documentclass{article}
\usepackage{spconf,amsmath,graphicx}

\usepackage{cite}
\usepackage{amsmath,amssymb,amsfonts}
\usepackage{algorithmic}
\usepackage{graphicx}
\usepackage{subcaption}
\usepackage{caption}
\usepackage{textcomp}
\usepackage{xcolor}
\usepackage{hyperref}

\title{Support Vector Data Description for Radar Target Detection}
\name{   J. Pinsolle$^{1}$\thanks{Part of this work was supported by ANR-ASTRID NEPTUNE 3 (ANR-23-ASM2-0009).},
        Y. A. Rouzoumka$^{1,2}$,
        C. Ren$^{1}$,
        C. Morisseau$^{2}$,
        J.-P. Ovarlez$^{1,2}$}
\address{$^{1}$SONDRA, CentraleSupélec, Université Paris-Saclay, 91190 Gif-sur-Yvette, France. \\$^{2}$  DEMR, ONERA, Université Paris Saclay, F-91123 Palaiseau, France}

\begin{document}

\maketitle

\begin{abstract}
Classical radar detection techniques rely on adaptive detectors that estimate the noise covariance matrix from target-free secondary data. While effective in Gaussian environments, these methods degrade in the presence of clutter, which is better modeled by heavy-tailed distributions such as the Complex Elliptically Symmetric (CES) and Compound-Gaussian (CGD) families. Robust covariance estimators like M-estimators or Tyler’s estimator address this issue, but still struggle when thermal noise combines with clutter. To overcome these challenges, we investigate the use of Support Vector Data Description (SVDD) and its deep extension, Deep SVDD, for target detection. These one-class learning methods avoid direct noise covariance estimation and are adapted here as CFAR detectors. We propose two novel SVDD-based detection algorithms and demonstrate their effectiveness on simulated radar data.
\end{abstract}


\section{Introduction}

Classical works on target detection proposed adaptive detection schemes that require an estimate of the noise covariance matrix, generally obtained from signal-free data traditionally called secondary or reference data. Early approaches assumed a Gaussian noise covariance structure, leading to detectors such as the Adaptive Matched Filter (AMF), where the noise is estimated using the sample covariance matrix (SCM) \cite{anmf, kelly, cfar}. However, such estimation performs poorly in non-Gaussian environments, typically when clutter is present, for instance, from forested areas or sea waves. Several statistical models have therefore been proposed to describe clutter, among which the Complex Elliptically Symmetric (CES) distribution and, more specifically, the Compound-Gaussian Distribution (CGD) account for most observed scenarios. Within this framework, $M$-estimators \cite{kelly} and Tyler’s estimator \cite{tyler, conv, asympanmf} present alternatives to the Gaussian-based SCM. Nevertheless, these approaches still face limitations in radar detection, particularly in the presence of additive white Gaussian thermal noise combined with clutter.


The target detection problem can be seen as a particular case of anomaly detection (AD), which has been extensively studied in the machine learning and deep learning literature \cite{review}.  In classical machine learning, one-class classification methods are among the most popular approaches to AD. Most of these models are inspired by Support Vector Machines, the most widely used being the one-class SVM \cite{svm, one_svm} and Support Vector Data Description (SVDD) \cite{svdd}. On the deep learning side, Deep SVDD \cite{dsvdd} extends the SVDD algorithm and represents a fully deep approach, training a network to map data into a minimum-volume hypersphere and thus directly addressing AD. In contrast, most deep AD methods rely on reconstruction error, in particular autoencoder-based methods \cite{ae}. Autoencoders learn compressed representations and reconstruct inputs, yielding low errors for normal and high errors for anomalous samples. 
Deep SVDD overcomes this by enforcing compactness through hypersphere minimization. Similarly, AnoGAN \cite{gan} leverages GANs to map test data into latent space and measure reconstruction error, but faces similar challenges in enforcing compact representations.\\
\indent To sum up, SVDD and Deep SVDD appear to be the most suitable machine learning and deep learning approaches for addressing Gaussian thermal noise combined with clutter scenarios in target detection. The remainder of this paper is organized as follows: Section \ref{sec:problem} presents the target detection problem and classical detectors; Section \ref{sec:SVDD} introduces two novel target detection algorithms based on SVDD and Deep SVDD; and Section \ref{sec:results} compares these methods with state-of-the-art approaches on simulated data.

\textit{Notations}: Matrices are in bold and capital, vectors in bold. For any matrix $\mathbf{A}$ or vector, $\mathbf{A}^H$ is the Hermitian transpose of $\mathbf{A}$. $\mathcal{CN}(\boldsymbol{\mu},\boldsymbol{\Gamma})$ is a complex circular Normal distribution of mean $\boldsymbol{\mu}$ and covariance matrix $\boldsymbol{\Gamma}$. The matrix operator $\boldsymbol{\mathcal{T}}(.)$ is the Toeplitz matrix operator $\rho \rightarrow \left\{\boldsymbol{\mathcal{T}}(\rho)\right\}_{i,j} = \rho^{|i-j|}$. $\left\| \cdot\right\|_F$ denotes the Frobenius norm.

\section{The detection problem and classical approaches to solve it}
\label{sec:problem}

Considering observed data $\mathbf{z} \in \mathbb{C}^m$, we aim to detect signals $\alpha \mathbf{p}$ corrupted by clutter and thermal noise. It can be stated as the following test:
\begin{align}
    &H_0: \mathbf{z} = \mathbf{c} + \mathbf{n} \, \nonumber ,\\
    &H_1: \mathbf{z} = \alpha \, \mathbf{p} + \mathbf{c} + \mathbf{n}\, .
\end{align}
 The vector $\mathbf{p}$ is known and called the \textit{steering vector}, $\alpha$ is the unknown complex target amplitude. $\mathbf{n}$ represents the thermal noise vector, which is considered white Gaussian. $\mathbf{c}$ represents the clutter noise. We consider two clutter models: Gaussian and compound Gaussian. In classical detection, we aim to estimate the clutter noise using secondary data $(\mathbf{z}_k)_{1 \leqslant k \leqslant K}$, which are assumed to be independent but depend on the clutter nature. In general, when no parameters are known, the Generalized Likelihood Ratio $\Lambda(\mathbf{z})$ is compared against a detection threshold $\lambda$:
\begin{equation}
    \Lambda(\mathbf{z}) = \frac{\displaystyle\max_{\theta, \mu} \; p_{\mathbf{z}/H_1}(\mathbf{z}, \theta, \mu)}{\displaystyle\max_{\mu} \; p_{\mathbf{z}/H_0}(\mathbf{z}, \mu)}
\;\; \underset{H_0}{\overset{H_1}{\gtrless}} \;\; \lambda\,,
\end{equation}
where $\theta$ denotes generic target parameters and $\mu$ denotes generic noise parameters. In the Gaussian case, the clutter $\mathbf{c}$ and the secondary data follow a complex circular Gaussian distribution with covariance matrix $\boldsymbol{\Sigma}$ denoted $\mathcal{C}\mathcal{N}(\mathbf{0}, \boldsymbol{\Sigma})$. In case $\boldsymbol{\Sigma}$ is known, and recalling here $\mathbf{p}$ is known but not $\alpha$, the log-likelihood ratio test leads to the so-called \textit{Matched Filter} (MF) given by:
 \begin{equation}
\Lambda_\mathrm{MF}(\mathbf{z}) = 
 \frac{\left|\mathbf{p}^H \, \boldsymbol{\Sigma}^{-1} \, \mathbf{z}\right|^2}{\mathbf{p}^H \, \boldsymbol{\Sigma}^{-1} \, \mathbf{p}}
\;\; \underset{H_0}{\overset{H_1}{\gtrless}} \;\; \lambda_\mathrm{MF}\, .
 \end{equation}
 Another case can be $\mathbf{c}$ following $\mathcal{C}\mathcal{N}(\mathbf{0}, \sigma^2 \boldsymbol{\Sigma})$ with known covariance matrix $\boldsymbol{\Sigma}$ but unknown variance $\sigma^2$, the Log-Likelihood Ratio test gives the \textit{Normalized Matched Filter} (NMF) \cite{anmf}:
\begin{equation}
 \Lambda_\mathrm{NMF}(\mathbf{z}) =
 \frac{\left|\mathbf{p}^H \, \boldsymbol{\Sigma}^{-1} \, \mathbf{z}\right|^2}{\left(\mathbf{p}^H \, \boldsymbol{\Sigma}^{-1} \, \mathbf{p}\right)\left(\mathbf{z}^H \, \boldsymbol{\Sigma}^{-1} \,\mathbf{z}\right)}
\;\; \underset{H_0}{\overset{H_1}{\gtrless}} \;\; \lambda_\mathrm{NMF}\, .
\end{equation}
In practice, the clutter covariance is often unknown and must be estimated from $K$ secondary data. In the Gaussian case, the covariance can be estimated by the Sample Covariance Matrix (SCM): $\hat{\mathbf{S}}_K = \displaystyle \frac{1}{K} \sum_{k=1}^K \mathbf{z}_k \, \mathbf{z}_k^H$. Plugging $\hat{\mathbf{S}}_K$ into the likelihood ratio yields the Adaptive Matched Filter (AMF) \cite{cfar}.
When the clutter follows $\mathcal{CN}(\mathbf{0}, \sigma^2 \boldsymbol{\Sigma})$ with unknown variance $\sigma^2$, one obtains the Adaptive Normalized Matched Filter (ANMF) \cite{anmfscm}.


In non-Gaussian environments, these detectors may degrade in terms of performance. A widely used model is the compound Gaussian representation, where clutter is expressed as $\mathbf{c} = \sqrt{\tau} \, \mathbf{g}$, with $\tau$ a positive random texture and $\mathbf{g}$ a zero-mean complex Gaussian vector with covariance $\boldsymbol{\Sigma}$, normalized such that $\mathrm{Tr}(\boldsymbol{\Sigma}) = m$ for identifiability.

To estimate $\boldsymbol{\Sigma}$ in this framework with $K$ secondary data, the Tyler estimator $\hat{\mathbf{M}}_{FP}$ \cite{tyler} is obtained by solving the fixed-point equation :
\begin{equation}
\hat{\mathbf{M}}_{FP} = \frac{m}{K} 
\sum_{k=1}^{K} 
\frac{\mathbf{z}_k \, \mathbf{z}_k^H}{\mathbf{z}_k^H \, \mathbf{M}_{FP}^{-1} \, \mathbf{z}_k} \, .
\end{equation}
Besides handling the compound Gaussian case, asymptotic studies show that Tyler and SCM estimators have slightly the same performances in Gaussian cases \cite{mahot}. Nevertheless, even with Tyler’s estimator, the resulting detectors struggle in more complex situations, particularly when white Gaussian thermal noise is present in addition to non-Gaussian clutter. This motivates the use of machine and deep learning methods, such as one-class classification algorithms, which rely on less restrictive assumptions and can thus provide powerful alternatives for the target detection problem.

\section{Presentation of One-class SVDD algorithms}
\label{sec:SVDD}
We now introduce SVDD and DSVDD algorithms and propose their adaptation to the target detection problem.

\subsection{SVDD}

Given the amplitudes of the complex signal, represented as feature vectors $\mathbf{z} \in \mathbb{C}^m$, the goal of SVDD is to find the smallest hypersphere enclosing the data in a Reproducing Kernel Hilbert Space (RKHS) $\mathcal{F}_k$. The mapping $\boldsymbol{\phi}_k : \mathbb{C}^m \to \mathcal{F}_k$ is defined implicitly through a positive definite kernel $k : \mathbb{C}^m\times\mathbb{C}^m \to \mathbb{R}$ such that $k(\mathbf{z}_1, \mathbf{z}_2) = \langle \boldsymbol{\phi}_k(\mathbf{z}_1), \boldsymbol{\phi}_k(\mathbf{z}_2) \rangle$, where $\langle \cdot , \cdot \rangle$ denotes the inner product in $\mathcal{F}_k$. The primal optimization problem seeks the center $\mathbf{c} \in \mathcal{F}_k$ and radius $R > 0$ of the hypersphere:
\begin{equation}
\label{eq:primal}
   \min_{R, \mathbf{c}, \xi} R^2 + \frac{1}{\nu N} \sum_{i=1}^N \xi_i\, ,
\end{equation}
subject to $\|\boldsymbol{\phi}_k(\mathbf{z}_i) - \mathbf{c}\|^2 \leq R^2 + \xi_i$, $\xi_i \geq 0$. The slack variables $\xi_i$ allow violations of the constraint, while $\nu$ controls the trade-off between the volume of the hypersphere and the fraction of tolerated outliers \cite{outlier}. The dual formulation is given by:
\begin{equation}
\label{eq:dual}
    \max_{\alpha} \quad \sum_i \alpha_i \, k(\mathbf{z}_i, \mathbf{z}_i) - \sum_{i,j} \alpha_i \,\alpha_j \,k(\mathbf{z}_i, \mathbf{z}_j)\, ,
\end{equation}
subject to $0 \leq \alpha_i \leq \displaystyle\frac{1}{\nu \,N}$ and $\displaystyle\sum_i \alpha_i = 1$. Solving this quadratic program yields the optimal multipliers $(\alpha_i^*)$, from which the decision function for a new point $\mathbf{z}$ can be computed as:
\begin{equation}
\label{eq:dist}
f(\mathbf{z}) = k(\mathbf{z},\mathbf{z}) - 2 \sum_i \alpha_i^* k(\mathbf{z}_i,\mathbf{z}) + \sum_{i,j} \alpha_i^* \alpha_j^* k(\mathbf{z}_i,\mathbf{z}_j).
\end{equation}

In target detection, maintaining a constant false alarm rate (CFAR) is crucial for reliable performance. Rather than using the estimated radius $R$, the detection decision is based on a threshold computed from secondary training data. Specifically, for a fixed false alarm probability $P_\text{fa}$, the distances of the secondary data to the SVDD center are evaluated using the distance function $f(\mathbf{z})$, and the threshold $\lambda_{SVDD}(P_\text{fa})$ is set as the $(1-P_\text{fa})$-quantile of these distances. 
A candidate $\mathbf{z}$ 
is declared a target if
\begin{equation}
\label{eq:condition}
f(\mathbf{z}) \;\; \underset{H_0}{\overset{H_1}{\gtrless}} \,  \lambda_{SVDD}(P_\text{fa}) \, .
\end{equation}

In the simulations, we will refer to this detector simply as SVDD. A known limitation of kernel-based SVDD is the high computational cost of solving the dual problem. This issue can be alleviated using neural network-based implementations, which approximate the feature mapping and significantly reduce computation.

\subsection{Deep SVDD}
Deep SVDD \cite{dsvdd} extends the classical SVDD approach to a neural network-based one-class classifier. Similar to SVDD, the goal is to map data into a hypersphere in a learned feature space $\mathcal{F}$. In Deep SVDD, the mapping function $\boldsymbol{\psi}(\cdot; \mathcal{W}): \mathbb{C}^m \rightarrow \mathcal{F}$ is implemented as a neural network with $L$ hidden layers and weights $\mathcal{W} = \{\mathbf{W}^1,...,\mathbf{W}^L\}$.

The network is trained to minimize the following loss:
\begin{equation}
\min_{\mathcal{W}} \quad \frac{1}{n}\sum_{i=1}^n \|\boldsymbol{\psi}(\mathbf{z}_i; \mathcal{W}) - \mathbf{c}\|^2
+ \frac{\beta}{2}\sum_{\ell=1}^L \left\|\mathbf{W}^\ell\right\|_F^2\, ,
\end{equation}
where the first term penalizes the squared distance of each network output from the hypersphere center $c$, and the second term is a weight decay regularization with hyperparameter $\beta$.

In contrast to the objective of classical SVDD \eqref{eq:primal}, which reduces the hypersphere radius and penalizes points lying outside it, this formulation contracts the hypersphere by minimizing the average distance of all representations to the center. Consequently, the network is forced to learn the shared factors of variation so that data points are mapped as close as possible to $\mathbf{c}$. By enforcing proximity for every sample, rather than tolerating outliers outside the sphere, the method aligns with the assumption that all training data belong to the same class. After training, we denote the network weights as $\mathcal{W}^*$ and the distance of a candidate $\mathbf{z}$ to the center as $||\boldsymbol{\psi}(\mathbf{z}; \mathcal{W}^*) - \mathbf{c}||^2$. A detection threshold $\lambda_{DSVDD}(P_\text{fa})$ is computed using a secondary training set, similarly to SVDD. The decision rule  for candidate $\mathbf{z}_i$ is then:
\begin{equation}
\label{eq:condition_dsvdd}
\|\boldsymbol{\psi}(\mathbf{z}_i; \mathcal{W}^*) - \mathbf{c}\|^2 \;\; \underset{H_0}{\overset{H_1}{\gtrless}} \,  \lambda_{DSVDD}(P_\text{fa}) \, .
\end{equation}
This formulation ensures a CFAR-like behavior while benefiting from the representation learning capability of neural networks, reducing computational complexity compared to kernel SVDD for large datasets.

\section{Simulations and Results}
\label{sec:results}

We evaluate the algorithms introduced in the previous sections, namely ANMF built with Tyler estimate, AMF-SCM, SVDD, and DSVDD. For reference, we also report the performance of the Matched Filter (MF) using the \emph{true} covariance matrix.

All simulations are conducted on synthetic radar-like data. Each dataset consists of $N$ independent samples, each sample being a matrix $\mathbf{Z} \in \mathbb{C}^{m \times K}$ formed by $K$ complex-valued column vectors of size $m$.
A stationary Toeplitz covariance matrix $\mathbf{\Sigma}=\boldsymbol{\mathcal{T}}(\rho)$ of size $m \times m$ with $\rho = 0.5$ is generated, and $K$ complex Gaussian clutter vectors $\mathbf{c} \sim \mathcal{CN}(\mathbf{0},\mathbf{\Sigma})$ are drawn for each sample.
If needed, a texture $\tau$ is drawn from a Gamma distribution $\Gamma(\mu, 1/\mu)$ ($\mu$ = 1), and each clutter vector is scaled by the texture $\sqrt{\tau}$ representing the power fluctuation. Additive white Gaussian thermal noise is then added. When a target is present, it is added as
    $\mathbf{z} = \alpha \, \mathbf{p} + \sqrt{\tau}\mathbf{c} + \mathbf{n} $ , with 
    $\alpha = \sqrt{\frac{\mathrm{SNR}}{m}} \, e^{j\phi}$, $\phi \sim \mathcal{U}[0,2\pi)$ and 
    $\mathbf{p} = \left(1, e^{j2\pi d/m}, \dots, e^{j2\pi d (m-1)/m}\right)^T$, 
    where $d$ is the Doppler bin index and $m=16$.

For the evaluation protocol, according to this process, several datasets are generated to train and evaluate each algorithm, following a false detection rate $P_{fa} = 0.01$. For SVDD and DSVDD algorithms, a set of $N = 5000$ training samples without a target is generated ($m = 16$, $K = 1$). For all algorithms, a set of $N$ data to determine the Pfa-threshold relationship is also generated without a target ($m=16$, $K=1$). For classical detectors, two sets of $N$ reference data are generated for the Pfa-threshold relationship determination and the test data ($m=16$, $K=32$).  $ N$ test data are generated with a target ($m = 16$, $K = 1$). They are the same for all algorithms.

\subsection{Parameters and Deep SVDD architecture}

The network consists of three successive 1D convolutional layers with 32, 64, and 128 filters, each followed by batch normalization, leaky ReLU activation, and max-pooling. An adaptive average pooling layer ensures a fixed-size output independent of the input length. Finally, a fully connected layer projects the features into a 128-dimensional representation. 

Complex-valued inputs are handled by concatenating their real and imaginary parts along the channel dimension. The network is trained for 15 epochs with a batch size of 64 on standardized input data. Optimization is completed via Adam optimizer \cite{adam} and uses a learning rate of 0.001, reduced by a factor $\gamma = 0.1$ at epoch 5 and 10 using a milestone scheduler. For the Deep SVDD objective, we apply $L_2$ weight decay regularization with coefficient $\beta= 0.001$.

The classical SVDD was implemented using a Gaussian (RBF) kernel with its width set to the inverse of the training data variance, and the loss parameter $\nu$ fixed to $0.01$.

\begin{figure}[htbp]
    \centering
    \begin{subfigure}[t]{0.48\textwidth}
        \centering
        \includegraphics[width=\linewidth]{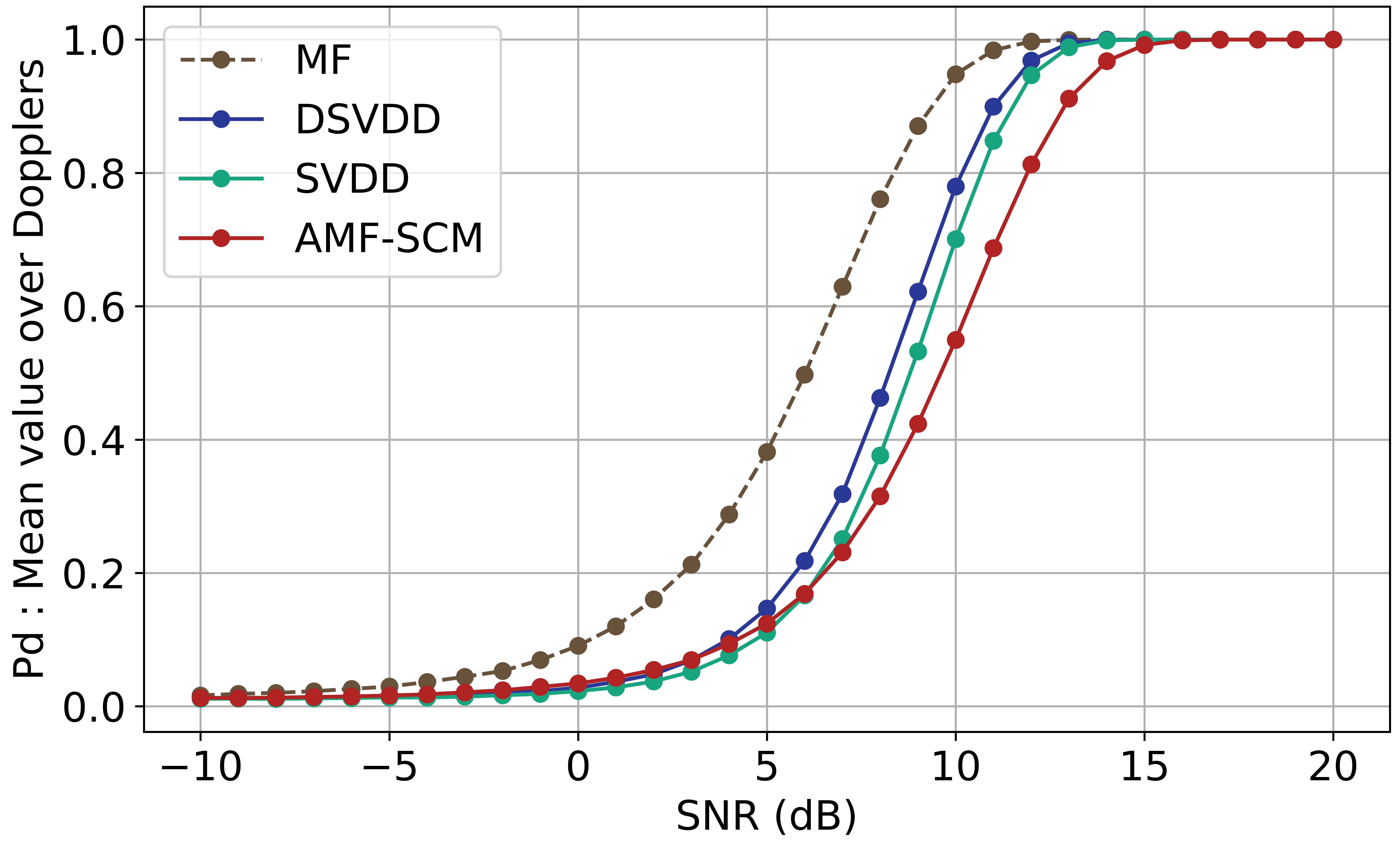}
        \caption{Gaussian + white Gaussian noise}
    \end{subfigure}
    \begin{subfigure}[t]{0.48\textwidth}
        \centering
        \includegraphics[width=\linewidth]{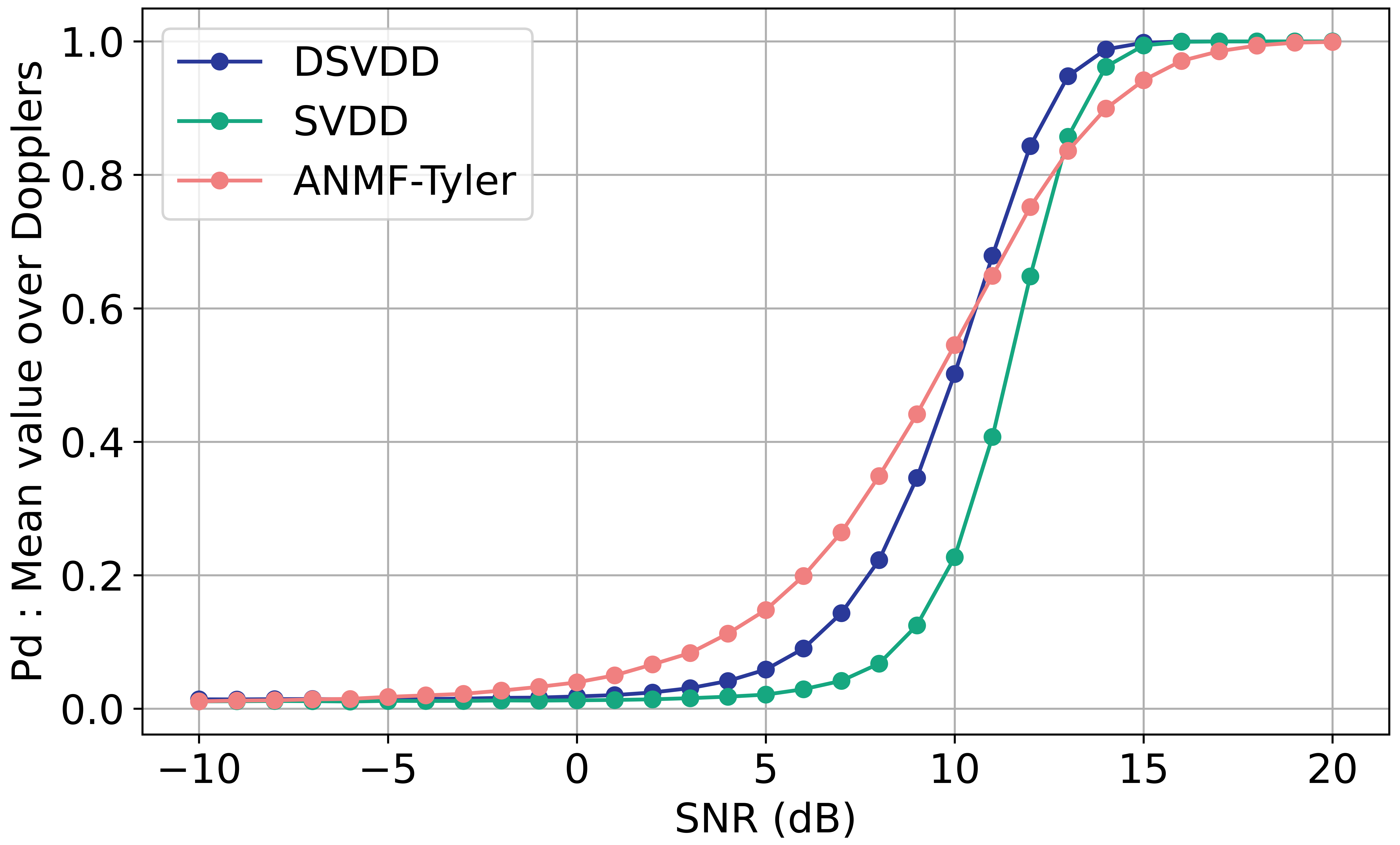}
        \caption{Compound Gaussian + white Gaussian noise}
    \end{subfigure}
    \caption{Detection performance of SVDD and Deep SVDD compared with classical detectors over 16 Doppler bins under $P_{fa} = 0.01$. Each curve shows the mean detection score across bins for SNR values from $-10\,$dB to $20\,$dB. In the Gaussian clutter scenario (a), the reference is MF, with AMF-SCM as the selected adaptive detector. For the Compound-Gaussian clutter with additive white noise (b), ANMF built with Tyler's estimate (ANMF-Tyler) is the chosen classical detector.}
    \label{fig:comparaison}
\end{figure}

\begin{figure}[htbp]
    \centering
    \includegraphics[width=1\linewidth]{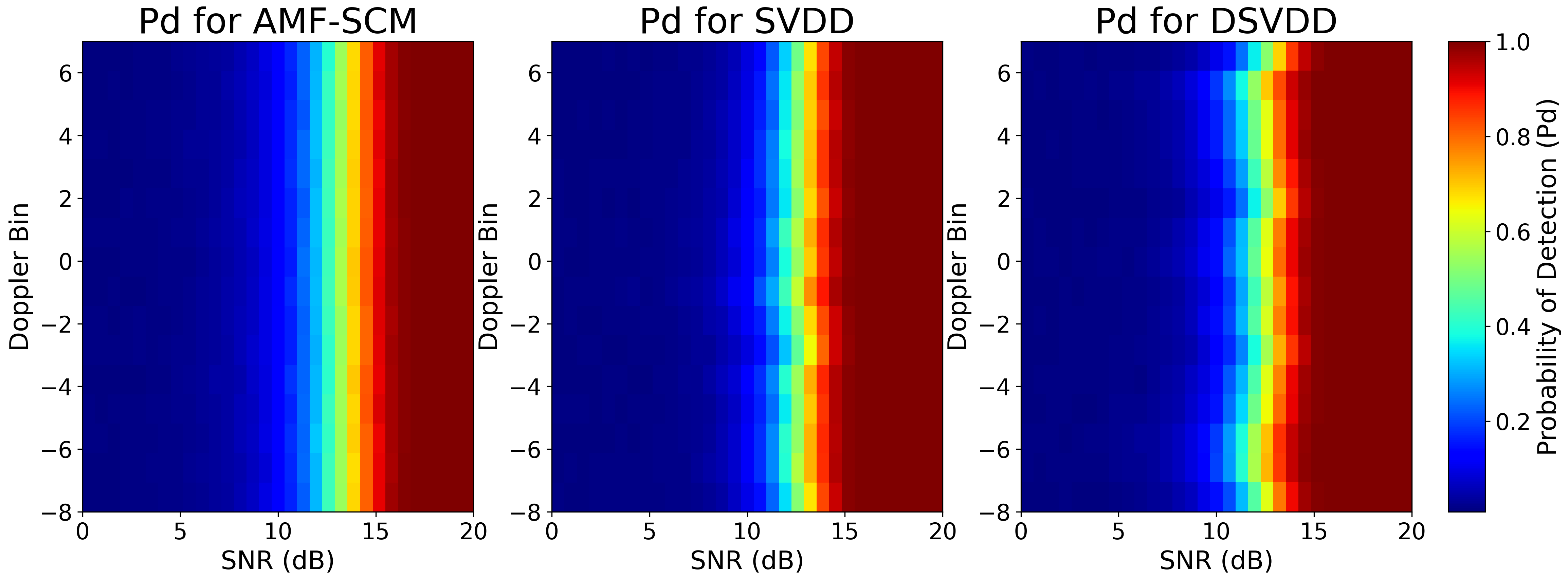}
    \includegraphics[width=1\linewidth]{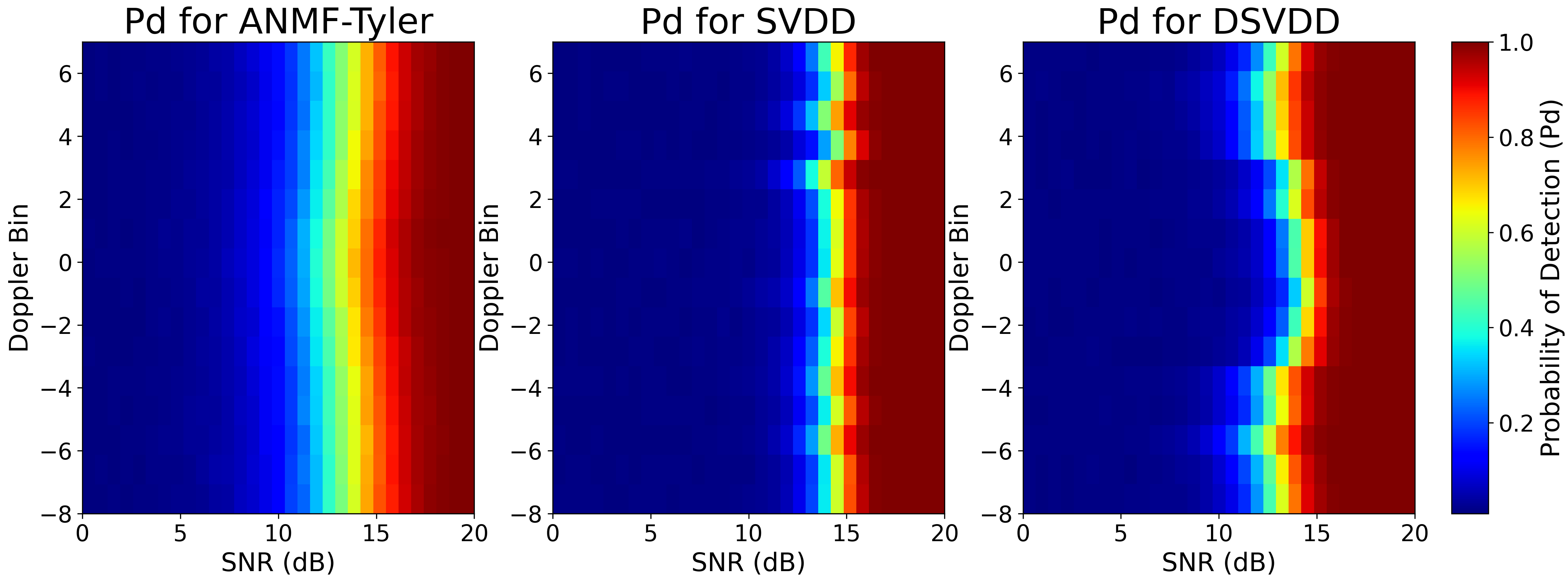}
 \caption{Map of the probability of detection (Pd) for each of the 16 Doppler bins, with SNR values ranging from 0 to 20 dB, for the considered detectors ($P_{fa} = 0.01$). Top plots: Gaussian clutter. Bottom plots: compound Gaussian clutter.}
\label{fig:map}
\end{figure}

\vspace{-0.3cm}
\subsection{Results}

Figure \ref{fig:comparaison} shows the probability of detection (Pd) as a function of the SNR, averaged over all Doppler bins. SVDD and DSVDD are compared to the AMF-SCM detector in the Gaussian clutter with additive white noise scenario, with the MF used as a reference. In the non-Gaussian cases with Compound Gaussian clutter with additive thermal noise, ANMF-Tyler is used as the main benchmark. Both SVDD-based detectors outperform AMF-SCM in the Gaussian scenario, especially at high SNRs. 
In the non-Gaussian scenarios with Compound-Gaussian clutter and additive thermal noise, the SVDD-based methods — particularly DSVDD — begin to outperform the ANMF-Tyler detector from approximately $11\,$dB onwards, despite being less effective at lower SNRs where ANMF-Tyler still achieves reasonable detection performance. This improvement is especially noteworthy since operational radar systems typically operate in the 10 dB to 15 dB SNR range.

To further analyze these results, Figure \ref{fig:map} presents the detection performance across all Doppler bins. In the Gaussian scenario with thermal noise, SVDD and DSVDD consistently outperform the other detectors across all bins. In the Compound-Gaussian scenario, DSVDD shows degraded performance around the zero-Doppler bin — known to be the most challenging case — yet still maintains strong performance for SNRs above 11 dB. SVDD exhibits similar behavior but lags behind its deep counterpart. ANMF-Tyler, on the other hand, provides stable performance across all bins and a wide SNR range.

\vspace{-0.3cm}
\section{Conclusion}
\label{sec:conclusion}

This work introduces two novel target detectors inspired by the one-class anomaly detection algorithms SVDD and DSVDD. Designed as CFAR detectors, they allow fair comparison with classical approaches from the target detection literature. Simulations show promising performance, with clear improvements over traditional detectors in scenarios involving clutter and additive thermal noise. However, these detectors still show weaknesses in Compound-Gaussian environments and around the zero-Doppler bin, both known as challenging cases. Further analysis of the latent space and experiments on real radar data could help enhance their robustness and better assess their practical potential.

\bibliographystyle{IEEEbib}
\bibliography{strings,refs}

\end{document}